\documentclass[review]{elsarticle}

\usepackage{lineno,hyperref}
\usepackage{graphicx}
\usepackage{times}
\usepackage{epsfig}
\usepackage{amsmath}
\usepackage{amssymb}
\usepackage{multirow}
\usepackage{color}
\usepackage{url}
\modulolinenumbers[5]

\journal{Pattern Recognition}

\begin{document}

\newcommand{\tabincell}[2]{
\begin{tabular}{@{}#1@{}}#2\end{tabular}
}

\begin{frontmatter}

\title{Center and Scale Prediction: Anchor-free Approach for Pedestrian and Face Detection}


\author{Wei Liu\textsuperscript{1*}}
\author{Irtiza Hasan\textsuperscript{2*}}
\cortext[mycorrespondingauthor]{Equal contribution}
\author{Shengcai Liao\corref{mycorrespondingauthor}\textsuperscript{2}}
\cortext[mycorrespondingauthor]{Corresponding author}

\address{National University of Defense Technology, Changsha, China \textsuperscript{1}}
\address{liuwei16@nudt.edu.cn}
\address{Inception Institute of Artificial Intelligence (IIAI), UAE \textsuperscript{2}}
\address{\{irtiza.hasan,shengcai.liao\}@inceptioniai.org}

\begin{abstract}
Object detection traditionally requires sliding-window classifier in modern deep learning based approaches. However, both of these approaches requires tedious configurations in bounding boxes. Generally speaking, single-class object detection is to tell where the object is, and how big it is. Traditional methods combine the "where" and "how" subproblems into a single one through the overall judgement of various scales of bounding boxes. In view of this, we are interesting in whether the "where" and "how" subproblems can be separated into two independent subtasks to ease the problem definition and the difficulty of training. Accordingly, we provide a new perspective where detecting objects is approached as a high-level semantic feature detection task. Like edges, corners, blobs and other feature detectors, the proposed detector scans for feature points all over the image, for which the convolution is naturally suited. However, unlike these traditional low-level features, the proposed detector goes for a higher-level abstraction, that is, we are looking for central points where there are objects, and modern deep models are already capable of such a high-level semantic abstraction. Like blob detection, we also predict the scales of the central points, which is also a straightforward convolution. Therefore, in this paper, pedestrian and face detection is simplified as a straightforward center and scale prediction task through convolutions. This way, the proposed method enjoys an anchor-free setting, considerably reducing the difficulty in training configuration and hyper-parameter optimization. Though structurally simple, it presents competitive accuracy on several challenging benchmarks, including pedestrian detection and face detection. Furthermore, a cross-dataset evaluation is performed, demonstrating a superior generalization ability of the proposed method.
\end{abstract}


\end{frontmatter}

\section{Introduction}
\label{intro}

Pedestrian detection is a very active research topic in the computer vision. It has applications in many different domains, including autonomous driving \cite{campmany2016gpu}, video surveillance \cite{hattori2015learning}, and action recognition \cite{zhang2020semantics}. Pedestrian detection also plays a critical role as foundation steps in various other computer vision research areas, such as multi-object tracking \cite{huang2019bridging}, human pose estimation \cite{wang2020combining}, person re-identification \cite{
zheng2017person, yan2021anchor}, etc. Recently, convolutional neural network (CNNs) based approaches have advanced the field of pedestrian detection a lot, with significantly improved performance.

For a broader domain, pedestrian detection is a special case of general object detection. Starting from the pioneering work of the Viola-Jones detector \cite{viola2004robust}, object detection generally requires sliding-window classifiers in traditional or anchor based predictions in CNN-based methods. These detectors are essentially local classifiers used to judge the pre-defined anchors (windows or anchor-boxes) as being objects or not. However, either of these approaches requires tedious configurations in anchors. Several (but not all) of these configurations, which include the number of scales, the sizes of anchors, the aspect ratios, and the overlap thresholds with ground truth boxes. All of these are task-oriented, and it is difficult to figure out which combination is the optimal one. Generally speaking, single-class object detection is to tell where the object is, and how big it is. Traditional methods combine the "where" and "how" subproblems into a single one through the overall judgement of various scales of anchor boxes. In view of this, we are interesting in whether the ``where" and ``how" subproblems can be separated into two independent subtasks to ease the problem definition and the difficulty of training.

We resort to traditional feature detection. Feature detection is one of the most fundamental problems in computer vision. It is usually viewed as a low-level technique, with typical tasks including edge detection (e.g. Canny \cite{canny1986computational}, Sobel \cite{sobel1972camera}), corner (or interest point) detection (e.g. SUSAN \cite{smith1997susan}, FAST \cite{rosten2006machine}), and blob (or region of interest point) detection (e.g. LoG \cite{lindeberg2013scale}, DoG \cite{lowe2004distinctive}, MSER \cite{matas2004robust}). Feature detection is of vital importance to a variety of computer vision tasks ranging from image representation, image matching to 3D scene reconstruction. Generally speaking, a feature is defined as an ``interesting part'' of an image. Therefore, feature detection aims to compute abstractions of image information. Subsequently,  feature detection makes local decisions at every image point whether there is an image feature of a given type at that point or not. With the rapid development for computer vision tasks, deep convolutional neural networks (CNN) are believed to be of very good capability to learn high-level image abstractions. CNNs have also been applied for feature detection, and demonstrate attractive successes even in low-level feature detections. For example, there is a recent trend of using CNN to perform edge detection \cite{shen2015deepcontour,xie2017holistically,bertasius2015deepedge,liu2017richer}, which has substantially advanced this field. It shows that clean and continuous edges can be obtained by deep convolutions, which indicates that CNN has a stronger capability to learn higher-level abstraction of natural images than traditional methods. This capability may not be limited to low-level feature detection; it may open up many other possibilities of high-level feature detection.

\begin{figure}[t]
\begin{center}
\includegraphics[width=0.8\linewidth]{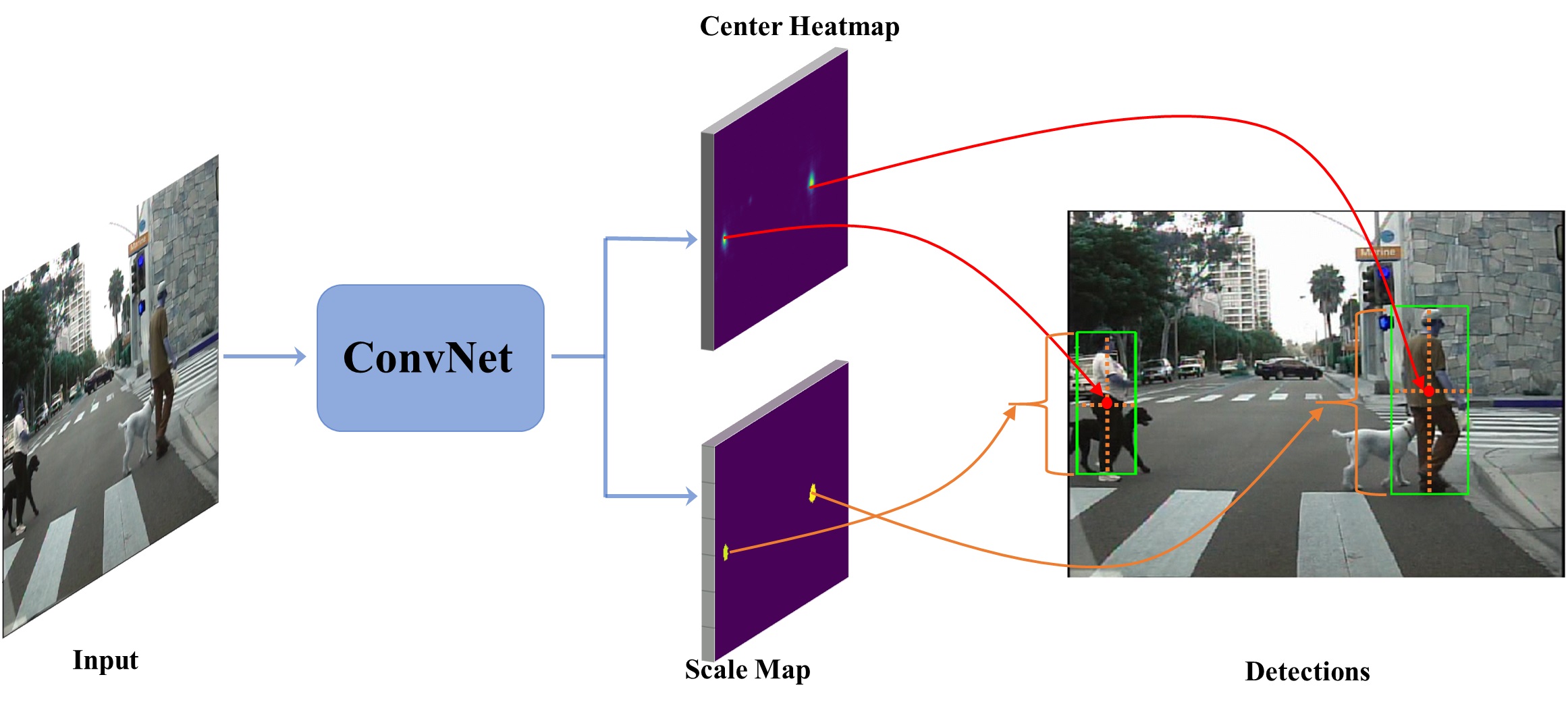}
\end{center}
   \caption{The overall pipeline of the proposed CSP detector. The final convolutions have two channels, one is a heatmap indicating the locations of the centers (red dots), and the other serves to predict the scales (yellow dotted lines) for each detected center.}
\label{fig:pipeline}
\end{figure}

Therefore, in this paper, we argue in the favor of posing pedestrian detection as high-level semantic feature detection task. However, unlike traditional low-level feature detectors, the proposed detector goes for a higher-level abstraction, that is, we are looking for central points where there are objects. Besides, similar to the blob detection, we also predict the scales of the central points. However, instead of processing an image pyramid to determine the scale as in traditional blob detection, we also predict object scale with a straightforward convolution in one pass upon a fully convolution network (FCN) \cite{long2015fully}, considering its strong capability. As a result, pedestrian and face detection is simply formulated as a straightforward center and scale prediction task via CNNs. Therefore, the proposed CSP detector separates the ``where" and ``how" subproblems into two different convolutions. These design choices enables CSP to enjoy an anchor-free (short for window-free or anchor-anchor-free) setting, considerably reducing the difficulty in training configuration and hyper-parameter optimization. The overall pipeline of the proposed method, denoted as \textbf{C}enter and \textbf{S}cale \textbf{P}rediction (\textbf{CSP}) based detector, is illustrated in Fig. \ref{fig:pipeline}.

Previously, FCN has already been applied to and made a success in multi-person pose estimation \cite{cao2016realtime,newell2017associative}, where several keypoints are firstly detected merely through responses of full convolutions, and then they are further grouped into complete poses of individual persons. In view of this, recently two inspirational works, CornerNet \cite{Law_2018_ECCV} and TLL \cite{Song_2018_ECCV}, successfully go free from windows and anchor-boxes, which perform object detection as convolutional keypoint detections and their associations. Though the keypoint association require additional computations, sometimes complex as in TLL, the keypoint prediction by FCN inspires us to go a step further, achieving center and scale prediction based pedestrian and face detection in full convolutions.

In summary, the main contributions of this work are as follows: (i) We show a new possibility that pedestrian and face detection can be simplified as a straightforward center and scale prediction task through convolutions, which bypasses the limitations of anchor-box based detectors and gets rid of the complex post-processing of recent keypoint pairing based detectors. (ii) The proposed CSP \cite{liu2019high} detector achieves the new state-of-the-art performance on two challenging pedestrian detection benchmarks, CityPersons \cite{zhang2017citypersons} and Caltech \cite{dollar2012pedestrian}, it also achieves competitive performance on one of the most popular face detection benchmark--WiderFace \cite{yang2016wider}.
(iii) The proposed CSP detector presents good generalization ability when cross-dataset evaluation is performed.

This work is built upon CSP \cite{liu2019high}. The major new content is additional experiments on face detection to demonstrate CSP's capability in detecting objects with various aspect ratios. Besides, we also conduct comparative experiments and analysis to demonstrate the generalization ability of the proposed detector. They are summarized as follows:
\begin{itemize}
\item  We evaluate the proposed method for face detection on one of the most popular face detection benchmarks, i.e. WiderFace \cite{yang2016wider}. The model is trained on the official training subset and evaluated on both the validation and test subsets. Comparable performance to other state-of-the-art face detectors on this benchmark are reported and thus demonstrates the proposed method's capability and competitiveness on face detection.
\item  To further evaluate the generalizability of the proposed CSP detector, we also conduct two cross-dataset evaluation experiments. For pedestrian detection, we compare the proposed CSP detector with the state-of-the-art anchor-box based pedestrian detector (ALFNet \cite{Liu_2018_ECCV}). Both of the two detectors are trained on the CityPersons \cite{zhang2017citypersons} training set and then are directly tested on Caltech \cite{dollar2012pedestrian}.
    For face detection, we compare the proposed CSP detector with the state-of-the-art anchor-box based face detector (DSFD \cite{li2018dsfd}). Both of the two detectors are trained on the WiderFace \cite{yang2016wider} training set and then are directly tested on FDDB \cite{fddbTech}, UCCS \cite{uccs} and DarkFace \cite{wei2018deep}. Experimental results show that the proposed CSP detector has a superior generalization ability than the compared methods.
\end{itemize}

Besides, we considerably reorganize the whole paper, especially the introduction part, for a better and more clear motivation.

\section{Related Works}
\subsection{Anchor-box based object detection}

Anchor based object detection is arguably the dominant paradigm in object detection, popularized by \cite{ren2015faster}. Architecturally speaking, one of the major component of anchor-box based object detectors are fixed anchor-boxes (pre-defined scales and aspect ratios).  Subsequently, these fixed anchor-boxes are used for classification and regression of final object bounding-box. Most two-stage based object detectors, such as Faster R-CNN \cite{ren2015faster}, generates proposals (in region proposal network) and progressively classifies and refines them in its downstream classifier (Fast R-CNN branch). Making it end-to-end trainable, in a single unified framework. In contrast to two stage methods, singe-stage detectors, such as SSD \cite{liu2016ssd}, without the proposal generation step and achieved comparable accuracy while are more efficient (computationally) than two-stage detectors.


Pedestrian detection can be broadly categorized into two sub-categories, 1) from the context of autonomous driving \cite{zhang2017citypersons} and 2) surveillance \cite{qian2019oriented}. In both cases, majority of the methods are direct extensions of Faster R-CNN. For example, \cite{zhang2016faster} was the first work to use RPN \cite{ren2015faster} along with random forest as a downstream classifier. MS-CNN \cite{cai2016unified} utilized the Faster R-CNN framework, but additionally exploited the multi-scale feature maps. Moreover, Zhang \emph{et al.} \cite{zhang2017citypersons} used standard faster R-CNN but with added strategies for training, such as high image resolution, larger receptive field, quantization for anchors etc. \cite{qian2019oriented} suggested a scale-aware Faster R-CNN. Despite achieving high accuracy on non-occluded pedestrians, occluded pedestrian detection stands as one of the major problems in pedestrian detection \cite{hasan2021generalizable}, initial works focused on tailoring of the loss function to address the problems \cite{wang2017repulsion,Zhang_2018_ECCV}. Subsequently, Bi-Box \cite{Zhou_2018_ECCV} extended Faster R-CNN by proposing a separate branch to predict the visible parts of the pedestrian. In addition to occlusion, small-scale pedestrians are also among the harder cases. \cite{zhang2018too} propose a reinforcement learning based approach to address small-scale pedestrians. Moreover,  \cite{zhao2019accurate} utilized pose estimation along with detection in a joint optimization for achieving better accuracy.
Unlike general object detection, single stage methods\cite{lee2020centermask,tian2019fcos} achieved superior performances than two stage methods in pedestrian detection. For instance, ALFNet \cite{Liu_2018_ECCV} and \cite{liu2019efficient} proposes the asymptotic localization fitting strategy to progressively refine anchor-boxes for accurate localization. Finally, \cite{Lin_2018_ECCV} ameliorated SSD architecture by focusing on the discriminative feature learning.

For face detection, it is dominated by the single-stage framework in recent years. Most of the advanced face detectors focus on the anchor-box design \cite{yu2018anchor} and matching strategies, because faces in the wild exhibits a large variation in size. Previously, \cite{yu2018anchor} proposed framework the relied on cascade of anchors at various scales to better detect faces of different sizes. Furthermore, FaceBoxes \cite{zhang2017faceboxes} introduces an anchor-box densification strategy to ensure anchor-boxes of different sizes have the same density on an image, and in \cite{zhang2017s3fd,zhu2018seeing}, the authors propose different anchor-box matching threshold to ensure a certain number of training examples for tiny faces, further DSFD \cite{li2018dsfd} proposes an improved anchor-box matching strategy to provide better initialization for the regressor. In SSH \cite{najibi2017ssh}, anchor-boxes with two neighboring sizes share the same detection feature map. PyramidBox \cite{Tang_2018_ECCV} designs novel PyramidAnchors to help contextual feature learning.

\subsection{Anchor-free object detection}
Anchor-free detectors bypass the requirement of anchor-boxes and detect objects directly from an image. DeNet \cite{tychsen2017denet} proposes to generate proposals by predict the confidence of each location belonging to four corners of objects. Following the two-stage pipeline, DeNet also appends another sub-network to re-score these proposals. Within the single-stage framework, YOLO \cite{redmon2016you} appends fully-connected layers to parse the final feature maps of a network into class confidence scores and box coordinates. Densebox \cite{huang2015densebox} devises a unified FCN that directly regresses the classification scores and distances to the boundary of a ground truth box on all pixels, and demonstrates improved performance with landmark localization via multi-task learning. Therefore, it is effective pipeline different from Faster R-CNN \cite{ren2015faster}. However, DenseBox resizes objects to a single scale during training, thus requiring image pyramids to detect objects of various sizes by multiple network passes during inference. Besides, Densebox defines the central area for each object and thus requires four parameters to measure the distances of each pixel in the central area to the object's boundaries. In contrast, the proposed CSP detector defines a single central point for each object, therefore two parameters measuring the object's scale are enough to get a bounding box, and sometimes one scale parameter is enough given uniform aspect ratio in the task of pedestrian detection.
Most recently, CornerNet \cite{Law_2018_ECCV} also applies a FCN but to predict objects' top-left and bottom-right corners and then group them via associative embedding \cite{newell2017associative}. Enhanced by the novel corner pooling layer, CornerNet achieves superior performance on MS COCO object detection benchmark \cite{lin2014microsoft}. Similarly, TLL \cite{Song_2018_ECCV} proposes to detect an object by predicting the top and bottom vertexes. To group these paired keypoints into individual instances, it also predicts the link edge between them and employs a post-processing scheme based on Markov Random Field. Applying on pedestrian detection, TLL achieves significant improvement on Caltech \cite{dollar2012pedestrian}, especially for small-scale pedestrians.

The proposed methodology is an an anchor-free detection method. However, our work is different from the conventional approaches, as such, that we try to illustrate that a single FCN can be effectively deployed for face and pedestrian detection. In doing so, we also show that a single center point can be used for precise localization without any post-processing methodologies (except NMS).    


\subsection{Feature detection}

Feature detection is a widely studied problem with extensive literature in computer vision. Broadly, it primarily includes edge detection \cite{canny1986computational,sobel1972camera}, corner detection \cite{rosten2006machine,rosten2010faster}, blob detection \cite{matas2004robust,deng2007principal} etc. Classical methods \cite{canny1986computational,sobel1972camera} of feature detection mostly relied on local cues, such as the brightness, colors, gradients and textures. However, after the emergence of CNN-based methods, the task of feature detection was greatly advanced. For instance, recently several methods deployed CNN-based methods to perform edge detection, such as, \cite{shen2015deepcontour,xie2017holistically,bertasius2015deepedge,liu2017richer}, which have substantially advanced this field. However, unlike the above mentioned methods that perform low-level feature detection (edge, corners and blobs), the proposed method aims for a higher level of abstraction, that is, our center of attention is localizing central points, where there are pedestrians, for which modern deep models are naturally suited. 




\section{Proposed Method}
\subsection{Preliminary}
Generally, with an input image $I$, the detection network may generate several feature maps with different resolutions, which can be defined as follows:
\begin{equation}
\phi_{i}=f_{i}(\phi_{i-1})=f_{i}(f_{i-1}(...f_{2}(f_{1}(I)))),
\label{eq1}
\end{equation}
where $\phi_{i}$ represents feature maps output by the $i$th stage. These feature maps are generated by $f_{i}(.)$, decreasing in size progressively. Given a network with $N$ downsampling stages, all the generated feature maps can be denoted as $\Phi=\{\phi_{1},\phi_{2},...,\phi_{N}\}$, upon which detection heads are further built.

We denote these feature maps that are responsible for detection as $\Phi_{det}$. Specifically, in a object detector, the feature maps responsible for detection can be represented as $\Phi_{det}=\{\phi_{L},\phi_{L+1},...,\phi_{N}\}$, where $1<L<N$. Besides $\Phi_{det}$, in anchor-based detectors, another key component is called anchors (termed as $\mathcal{B}$). Given $\Phi_{det}$ and $\mathcal{B}$ in hand, detection can be formulated as:
\begin{equation}
\begin{aligned}
Dets&=\mathcal{H}(\Phi_{det},\mathcal{B}) \\
&=\{cls(\Phi_{det},\mathcal{B}),regr(\Phi_{det},\mathcal{B})\},
\end{aligned}
\label{eq2}
\end{equation}
where $\mathcal{B}$ is pre-defined according to the corresponding set of feature maps $\Phi_{det}$, and $\mathcal{H}(.)$ represents the detection head. Generally, $\mathcal{H}(.)$ contains two elements, namely $cls(.)$ which predicts the classification scores, and $regr(.)$ which predicts the scaling and offsets of the anchors.

While in anchor-free detectors, detection is performed merely on the set of feature maps $\Phi_{det}$, that is,
\begin{equation}
Dets=\mathcal{H}(\Phi_{det})
\label{eq3}
\end{equation}

\begin{figure*}
\begin{center}
\includegraphics[width=1.0\linewidth]{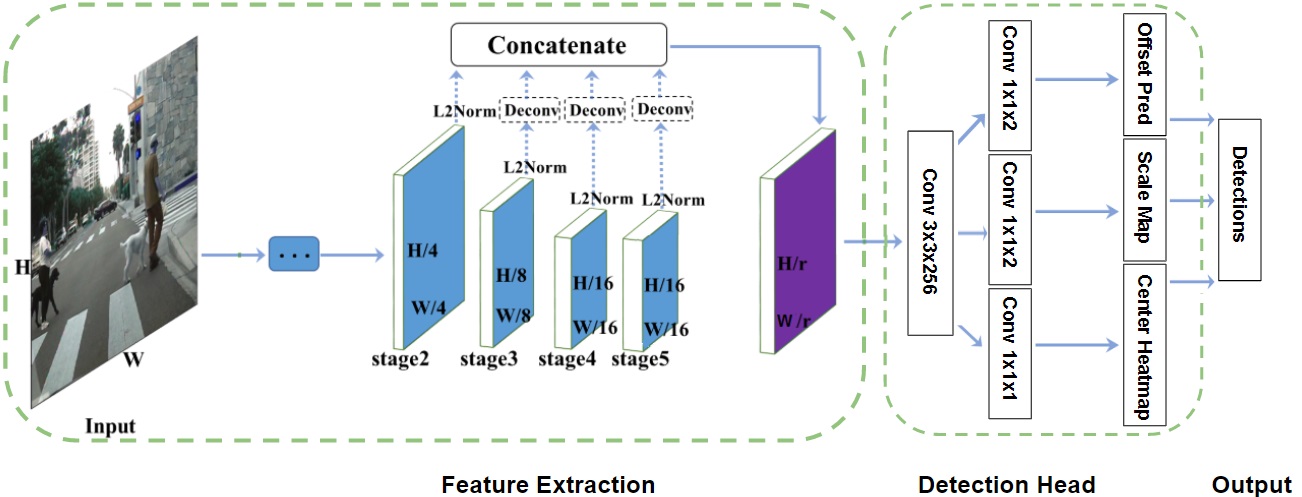}
\end{center}
   \caption{Overall architecture of CSP, which mainly comprises two components, i.e. the feature extraction module and the detection head. The feature extraction module concatenates feature maps of different resolutions into a single one. The detection head merely contains a 3x3 convolutional layer, followed by three prediction layers, for the offset prediction, center location and the the corresponding scale.}
\label{fig:arc}
\end{figure*}

\subsection{Overall architecture}\label{sec:arc}
The proposed CSP belongs to the group of anchor-free detectors, the overall architecture is illustrated in Fig. \ref{fig:arc}. The backbone network are truncated from a standard network pretrained on ImageNet \cite{deng2009imagenet}.

\textbf{Feature Extraction}.
Taking ResNet-50 as an example, it can be divided into five stages, where the output feature maps of each stage are downsampled by 2, 4, 8, 16, 32 w.r.t. the input image. As a common practice \cite{wang2017repulsion,Song_2018_ECCV}, the dilated convolutions are adopted in \emph{stage 5} to maintain the final feature map with a higher resolution. We denote the output of \emph{stage 2, 3, 4 and 5} as $\phi_{2}$, $\phi_{3}$, $\phi_{4}$ and $\phi_{5}$, in which the shallower feature maps can provide more precise localization information, while the coarser ones contain more semantic information with increasing the sizes of receptive fields. We simply fuse these multi-scale feature maps by concatenation, before which a deconvolution layer is adopted to make multi-scale feature maps with the same resolution. Since the feature maps from each stage have different scales, we use L2-normalization to rescale their norms to 10, as adopted in \cite{Lin_2018_ECCV}. To investigate the optimal combination from these multi-scale feature maps, we conduct an ablative experiment in Sec. \ref{sec:abl}. Given an input image of size $H \times W$, the size of final concatenated feature maps is $H/r \times W/r$, where $r$ is the downsampling factor. Similarly to \cite{Song_2018_ECCV}, $r=4$ gives the best performance as demonstrated in our experiments, because a larger $r$ means coarser feature maps which struggle on accurate localization. Given its simplicity, more complicated feature fusion strategies in \cite{lin2016feature,Kim_2018_ECCV,Kong_2018_ECCV} can be explored to further improve the performance, but it is not in the scope of this work.

\textbf{Detection Head}.
Upon the concatenated feature maps $\Phi_{det}$, a detection head is appended to parse it into detection results. As stated in \cite{rfbnet}, the detection head plays a significant role. In this work, we firstly attach a single 3x3 \emph{Conv} layer on $\Phi_{det}$ to reduce its channel dimensions to 256, and then two sibling 1x1\emph{Conv} layers are appended to produce the scale map and center heatmap, respectively. Also, we do this for simplicity and any improvement of the detection head  \cite{fu2017dssd,rfbnet,li2017light,li2018dsfd} can be flexibly incorporated into this work to be a better detector.

Optionally, to slightly adjust the center location, an extra offset prediction branch can be appended in parallel with the above two branches. This extra offset branch resolves mismatching of centers that occurs due to downsampled feature maps by learning the shifted offset. We will demonstrate the effectiveness of the extra offset branch in (Sec. \ref{sec:abl}).

\begin{figure}[t]
\begin{center}
\includegraphics[width=0.9\linewidth]{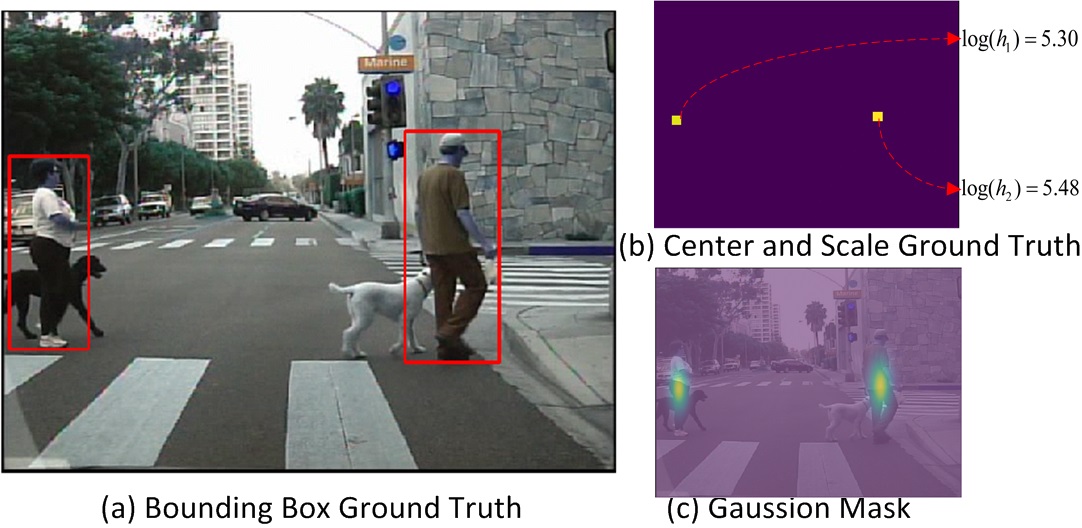}
\end{center}
   \caption{(a) is the bounding box annotations commonly adopted by anchor-box based detectors. (b) is the center and scale ground truth generated automatically from (a). Locations of all objects' center points are assigned as positives, and negatives otherwise. Each pixel is assigned a scale value of the corresponding object if it is a positive point, or 0 otherwise. We only show the height information of the two positives for clarity. (c) is the overall Gaussian mask map $M$ defined in Eq.\ref{eq4} to reduce the ambiguity of these negatives surrounding the positives.}
\label{fig:gt}
\end{figure}
\subsection{Training}
\textbf{Ground Truth}.
The predicted heatmaps are with the same size as the concatenated feature maps (i.e.  $H/r \times W/r$). An illustration example is depicted in Fig. \ref{fig:gt} (b) to illustrate how to generate the center and scale ground truth by the given bounding box annotations. For the center ground truth, the location where an object's center point falls in is assigned as positive while all others are negatives.

Scale can be defined as the height and/or width of objects.
In pedestrian detection, line annotation is first proposed in \cite{zhang2018towards,zhang2017citypersons} to generate high-quality ground truth, where tight bounding boxes are automatically generated with a uniform aspect ratio \footnote{Aspect ratio here is defined as width $/$ height} of 0.41. In accordance to this annotation, we can merely predict the height of each pedestrian instance and generate the bounding box with the predetermined aspect ratio. Therefore, for the scale ground truth, the $k$th positive location is assigned with the value of $log(h_{k})$. To reduce the center ambiguity, $log(h_{k})$ is also assigned to the negatives within a radius 2 of the positives, while all other locations are assigned as zeros.
Alternatively, we can also predict the width or height+width, which is required for face detection, because face objects in the wild exhibit a large variation in aspect ratios.

When the offset prediction branch is appended, the ground truth for the offsets of those centers can be defined as $(\frac{x_{k}}{r}-\lfloor \frac{x_{k}}{r} \rfloor, \frac{y_{k}}{r}-\lfloor \frac{y_{k}}{r} \rfloor)$.

\textbf{Loss Function.}
The center prediction can be formulated as a classification task. Note that it is difficult to decide an 'exact' center point. In order to reduce the ambiguity of these negatives surrounding the positives, we also apply a 2D Gaussian mask $G(.)$ centered at the location of each positive, which is also adopted in \cite{Law_2018_ECCV,Song_2018_ECCV}. An example of the overall mask map $M$ is depicted in Fig. \ref{fig:gt} (c). Formally, it is formulated as:
\begin{equation}
\begin{aligned}
&M_{ij}=\max_{k=1,2,...,K} G(i,j;x_{k},y_{k},\sigma_{w_{k}},\sigma_{h_{k}}), \\
&G(i,j;x,y,\sigma_{w},\sigma_{h})=e^{-(\frac{(i-x)^2}{2\sigma_{w}^2}+\frac{(j-y)^2}{2\sigma_{h}^2})},
\end{aligned}
\label{eq4}
\end{equation}
where $K$ is the number of objects in an image, $(x_{k},y_{k},w_{k},h_{k})$ is the center coordinates, width and height of the $k$th object, and the variances $(\sigma^{k}_{w},\sigma^{k}_{h})$ of the Gaussian mask are proportional to the height and width of individual objects. If these masks have overlaps, we choose the maximum values for the overlapped locations.
To combat the extreme positive-negative imbalance problem, the focal weights \cite{lin2017focal} on hard examples are also adopted.
Thus, the classification loss can be formulated as:
\begin{equation}
L_{center} = -\frac{1}{K}\sum_{i=1}^{W/r}\sum_{j=1}^{H/r}\alpha_{ij}(1-\hat{p}_{ij})^{\gamma}log(\hat{p}_{ij}),\\
\label{eq:clsloss}
\end{equation}
where

\begin{equation}
\begin{aligned}
&\hat{p}_{ij}=
\begin{cases}
p_{ij}  & \mbox{if $y_{ij}=1$}\\
1-p_{ij}& \mbox{otherwise,}
\end{cases} \\
&\alpha_{ij}=
\begin{cases}
1  & \mbox{if $y_{ij}=1$}\\
(1-M_{ij})^{\beta}& \mbox{otherwise.}
\end{cases}
\end{aligned}
\end{equation}

In the above, $p_{ij}\in[0,1]$ is the center prediction in the location $(i,j)$, and $y_{ij}\in\{0,1\}$ specifies the ground truth label, where $y_{ij}=1$ represents the positives and $y_{ij}=0$ the negatives. $\gamma$ is the focusing hyper-parameter, which is set as 2 as suggested in \cite{lin2017focal}. To reduce the ambiguity from those negatives surrounding the positives, the $\alpha_{ij}$ according to the Gaussian mask $M$ is applied to reduce their contributions to the total loss, in which the hyper-parameter $\beta$ controls the penalty. Experimentally, $\beta=4$ gives the best performance, which is similar in \cite{Law_2018_ECCV}.

The scale prediction can be formulated as a regression task via smooth L1 loss \cite{girshick2015fast}:
\begin{equation}
L_{scale} = \frac{1}{K}\sum_{k=1}^{K}SmoothL1(s_{k},t_{k}),
\label{eq:sml1}
\end{equation}

where $s_{k}$ and $t_{k}$ represents the network's prediction and the ground truth of each positive, respectively.

If the offset prediction branch is appended, the similar smooth L1 loss in Eq. \ref{eq:sml1} is adopted (denoted as $L_{offset}$).

To sum up, the full optimization objective is:
\begin{equation}
L = \lambda_{c} L_{\it{center}} + \lambda_{s} L_{\it{scale}} +\lambda_{o} L_{\it{offset}},
\label{eq:jointloss}
\end{equation}
where $\lambda_{c}$, $\lambda_{s}$ and $\lambda_{o}$ are the weights for center classification, scale regression and offset regression losses, which are experimentally set as 0.01, 1 and 0.1, respectively.

\subsection{Inference}
For inference, CSP simply involves a single forward of the network. Specifically, given the predictions, locations with confidence score above 0.01 in the center heatmap are kept, along with their predicted scale in the scale map. Then bounding boxes are generated and remapped to the original image size, followed by NMS. If the offset prediction branch is adopted, the centers are shifted according to the predicted offsets before being remapped to the original image size.

\section{Experiments}
\subsection{Experiment settings}
\subsubsection{Datasets and evaluation metrics}
We evaluate our approach on two tasks, including pedestrian detection and face detection.

For pedestrian detection, we choose two of the largest benchmark datasets: Caltech \cite{dollar2012pedestrian} and CityPersons \cite{zhang2017citypersons}.
The Caltech dataset contains approximately 2.5 hours of autodriving video with extensively labelled bounding boxes. Following \cite{zhang2017citypersons,mao2017can,wang2017repulsion,Liu_2018_ECCV,Zhang_2018_ECCV}, we use the training data augmented by 10 folds (42782 frames) and the standard test set with 4024 frames, evaluation results are reported based on the new annotations provided by \cite{zhang2016far}.
The CityPersons dataset is a more challenging benchmark with various occlusion levels. We train the models on the official training set with 2975 images and test on the validation set with 500 images.
Evaluation follows the standard Caltech evaluation metric \cite{dollar2012pedestrian}, that is log-average Miss Rate (MR) over False Positive Per Image (FPPI) ranging in [$10^{-2}$, $10^{0}$] (denoted as $MR^{-2}$).

For face detection, we choose one of the most challenging face detection benchmark, i.e. WiderFace \cite{yang2016wider}. The main reason we choose this dataset is due to its large variability of scale, aspect ratios, illumination and occlusions. The dataset defines three levels of difficulty: Easy, Medium and Hard by the detection rate of EdgeBox \cite{zitnick2014edge}. We train the proposed CSP merely on the training subset and test on both validation and testing subsets by the Average Precision (AP).

We first conduct the ablative study on the Caltech dataset, and then compare the proposed CSP detector with the state of the arts on all the above benchmarks.

\subsubsection{Training details}
The proposed method is implemented in Keras\footnote{https://github.com/fchollet/keras}.
We use ResNet-50 \cite{he2016deep} pretrained on ImageNet \cite{deng2009imagenet} as the backbone unless otherwise stated. Adam \cite{kingma2014adam} optimizer is applied and the strategy of moving average weights \cite{tarvainen2017mean} is used to achieve more stable training.
During training, standard data augmentations include random color distortion, horizontal flip, scaling and crop, the input is finally resized to 336x448, 640x1280 and 704x704 pixels for Caltech, CityPersons and WiderFace, respectively.

For Caltech, a mini-batch contains 16 images with one GPU (GTX 1080Ti), the learning rate is set as $10^{-4}$ and training is stopped after 15K iterations, the model initialized from CityPersons \cite{zhang2017citypersons} is trained with the learning rate of $2\times10^{-5}$.
For CityPersons, a mini-batch contains 8 images on 4 GPUs with the learning rate of $2\times10^{-4}$ and training is stopped after 37.5K iterations.
For WiderFace, a mini-batch contains 16 images on 8 GPUs with the learning rate of $2\times10^{-4}$ and training is stopped after 99.5K iterations. Similar data augmentations in PyramidBox \cite{Tang_2018_ECCV} is applied to increase the proportion of small faces during training.

\subsection{Ablation Study}\label{sec:abl}
We perform ablation study of the key components of CSP on the Caltech dataset.

\textbf{Why do we use the Center Point?}
The center point is capable of locating an individual object. A coming question is how about other high-level feature points. To answer this, we choose two other high-level feature points adopted in \cite{Song_2018_ECCV}, i.e. the top and bottom vertexes. Comparisons are reported in Table. \ref{table:point}. It is shown that both the two vertexes can succeed in detection but underperform the center point by approximately 2\%-3\% under IoU=0.5, and the performance gap is even larger under the stricter IoU=0.75. This is probably because the center point is advantageous to perceive the full body information.

\textbf{How important is the Scale Prediction?}
Scale prediction is another indespensible component of CSP. In practice, we merely predict the height for each detected center. To demonstrate the generality of CSP, we have also tried to predict Width or Height+Width for comparison. For Height+Width, the only difference lies in that the scale prediction branch has two channels responsible for the height and width respectively. It can be observed in Table \ref{table:scale} that Width and Height+Width prediction can also achieve comparable but suboptimal results to Height prediction. This result may be attributed to the line annotation style \cite{zhang2018towards,zhang2017citypersons} which provides accurate Height information, while Width is automatically generated by a fixed aspect ratio and thus is not able to provide additional information for training.
Anyway, CSP is potentially feasible for other detection tasks requiring both height and width, which will be demonstrated in the following experiments for face detection.
\begin{table}[t]
\begin{center}
\begin{tabular}{c|c|c}
\hline
\multirow{2}{*}{\tabincell{c}{Point\\Prediction}} & \multicolumn{2}{|c}{$MR^{-2}$(\%)}\\
\cline{2-3} {} & IoU=0.5 & IoU=0.75\\
\hline
\hline
Center point & \textbf{4.62}& \textbf{36.47} \\
\hline
Top vertex & 7.75& 44.70 \\
\hline
Bottom vertex & 6.52& 40.25 \\
\hline
\end{tabular}
\end{center}
\caption{Comparisons of different high-level feature points. Bold number indicates the best result.}
\label{table:point}
\end{table}

\begin{table}
\begin{center}
\begin{tabular}{c|c|c}
\hline
\multirow{2}{*}{\tabincell{c}{Scale\\Prediction}} & \multicolumn{2}{|c}{$MR^{-2}$(\%)}\\
\cline{2-3} {} & IoU=0.5 & IoU=0.75\\
\hline
\hline
Height & \textbf{4.62}& \textbf{36.47}\\
\hline
Width & 5.31& 53.06 \\
\hline
Height+Width & 4.73& 41.09\\
\hline
\end{tabular}
\end{center}
\caption{Comparisons of different definitions for scale prediction. Bold number indicates the best result.}
\label{table:scale}
\end{table}

\begin{table*}
\begin{center}
\begin{tabular}{c |c|c|c|c|c|c}
\hline
\multirow{2}{*}{\tabincell{c}{Feature for\\Detection}} & \multirow{2}{*}{+Offset} & \multirow{2}{*}{\tabincell{c}{Test Time\\(ms/img)}} & \multicolumn{2}{|c|}{$MR^{-2}$(\%)}&\multicolumn{2}{|c}{$\Delta MR^{-2}$(\%)}\\
\cline{4-7} {} & {} & {} & IoU=0.5 & IoU=0.75 & IoU=0.5 & IoU=0.75\\
\hline
\hline
$\Phi^{2}_{det}$ & {} & 69.8 & 5.32 & 30.08 & - & - \\
\hline
\multirow{2}{*}{$\Phi^{4}_{det}$} & {} & 58.2 & 4.62& 36.47 & \multirow{2}{*}{+0.08} &\multirow{2}{*}{+7.67}\\
\cline{2-5} {} & $\checkmark$ & 59.6 & \textbf{4.54}& \textbf{28.80} & {} & {}\\
\hline
\multirow{2}{*}{$\Phi^{8}_{det}$} & {} & 49.2 & 7.00& 54.25 & \multirow{2}{*}{+0.92} &\multirow{2}{*}{+21.32}\\
\cline{2-5} {} & $\checkmark$ & 50.4 & 6.08& 32.93 & {} & {}\\
\hline
\multirow{2}{*}{$\Phi^{16}_{det}$} & {} & 42.0 & 20.27 & 75.17 & \multirow{2}{*}{\textbf{+12.86}} &\multirow{2}{*}{\textbf{+41.30}}\\
\cline{2-5} {} & $\checkmark$ & 42.7 & 7.41 & 33.87 & {} & {}\\
\hline
\end{tabular}
\end{center}
\caption{Comparisons of different downsampling factors of the feature maps, which are denoted as $\Phi^{r}_{det}$ downsampled by $r$ w.r.t the input image. Test time is evaluated on the image with size of 480x640 pixels. $\Delta MR^{-2}$ means the improvement from the utilization of the offset prediction. Bold numbers indicate the best result.}
\label{table:down}
\end{table*}

\begin{table*}
\begin{center}
 \resizebox{.999\textwidth}{!}{
\begin{tabular}{c c c c|c c c|c c c}
\hline
\multicolumn{4}{c|}{Feature Maps} & \multicolumn{3}{|c|}{ResNet-50\cite{he2016deep}} &\multicolumn{3}{|c}{MobileNetV1\cite{howard2017mobilenets}} \\
\hline
$\phi_{2}$ & $\phi_{3}$ & $\phi_{4}$ & $\phi_{5}$ & \# Parameters & Test Time & $MR^{-2}$(\%) & \# Parameters & Test Time & $MR^{-2}$(\%) \\
\hline
\hline
$\checkmark$ & $\checkmark$ & {} & {} & 4.7MB & 36.2ms/img & 9.96 & 2.1MB & 27.3ms/img & 34.96\\
{} & $\checkmark$ & $\checkmark$ & {} & 16.1MB & 44.5ms/img & 5.68 & 6.0MB & 32.3ms/img & \textbf{8.33}\\
{} & {} & $\checkmark$ & $\checkmark$ & 37.4MB & 54.4ms/img & 5.84 & 10.7MB & 34.5ms/img & 10.03\\
$\checkmark$ & $\checkmark$ & $\checkmark$ & {} & 16.7MB & 46.0ms/img & 6.34 & 6.3MB & 33.3ms/img & 8.43\\
{} & $\checkmark$ & $\checkmark$ & $\checkmark$ & 40.0MB & 58.2ms/img & \textbf{4.62}& 12.3MB & 38.2ms/img & 9.59\\
$\checkmark$ & $\checkmark$ & $\checkmark$ & $\checkmark$ & 40.6MB & 61.1ms/img & 4.99& 12.6MB & 40.5ms/img & 9.05\\
\hline
\end{tabular}
}
\end{center}

\caption{Comparisons of different combinations of multi-scale feature representations defined in Sec. \ref{sec:arc}. $\phi_{2}$, $\phi_{3}$, $\phi_{4}$ and $\phi_{5}$ represent the output of \emph{stage 2, 3, 4 and 5} of a backbone network, respectively.  Bold numbers indicate the best results.}
\label{table:featcom}

\end{table*}

\textbf{How important is the Feature Resolution?}
In the proposed method, the final set of feature maps (denoted as $\Phi^{r}_{det}$) is downsampled by $r$ w.r.t the input image. To explore the influence from $r$, we train the models with $r=2,4,8,16$ respectively. For $r=2$, $\Phi^{2}_{det}$ are up-sampled from $\Phi^{4}_{det}$ by deconvolution. For $r=4,8,16$, the offset prediction branch is alternatively appended.
Stricter evaluations under IoU=0.75 are included to verify the effectiveness of additional offset prediction.
As can be seen from Table. \ref{table:down}, without offset prediction, $\Phi^{4}_{det}$ presents the best result under IoU=0.5, but performs poorly under IoU=0.75 when compared with $\Phi^{2}_{det}$, which indicates that finer feature maps are beneficial for precise localization. Not surprisingly, a larger $r$ witnesses a significant performance drop, which is mainly because coarser feature maps result in poor localization. In this case, additional offset prediction can substantially improve the detector upon $\Phi^{16}_{det}$ by 12.86\% and 41.30\% under the IoU threshold of 0.5 and 0.75, respectively. It also brings an improvement of 7.67\% under IoU=0.75 for the detector upon $\Phi^{4}_{det}$, with negligible extra computation cost, approximately 1ms per image of 480x640 pixels.

\textbf{How important is the Feature Combination?}
It is revealed in \cite{Song_2018_ECCV} that multi-scale representation is vital for detection of various scales. In this part, we consider different combinations of multi-scale feature maps from the backbone. In practice we choose the output of stage 2 ($\phi_{2}$) as a start point and the downsampling factor $r$ is fixed as 4. In spite of the ResNet-50~\cite{he2016deep} with \emph{stronger} feature representation, a light-weight network like MobileNetV1~\cite{howard2017mobilenets} is also choosen. Results in Table \ref{table:featcom} reveal that the much shallower feature maps like $\phi_{2}$ result in poorer accuracy, while deeper feature maps like $\phi_{4}$ and $\phi_{5}$ are of great importance for superior performance, and the middle-level feature maps $\phi_{3}$ are indispensable to achieve the best results. For ResNet-50, the best performance comes from the combination of $\{\phi_{3},\phi_{4},\phi_{5}\}$, while $\{\phi_{3},\phi_{4}\}$ is the optimal one for MobileNetV1.

\begin{figure*}[t]
\begin{center}
\includegraphics[width=1.0\linewidth]{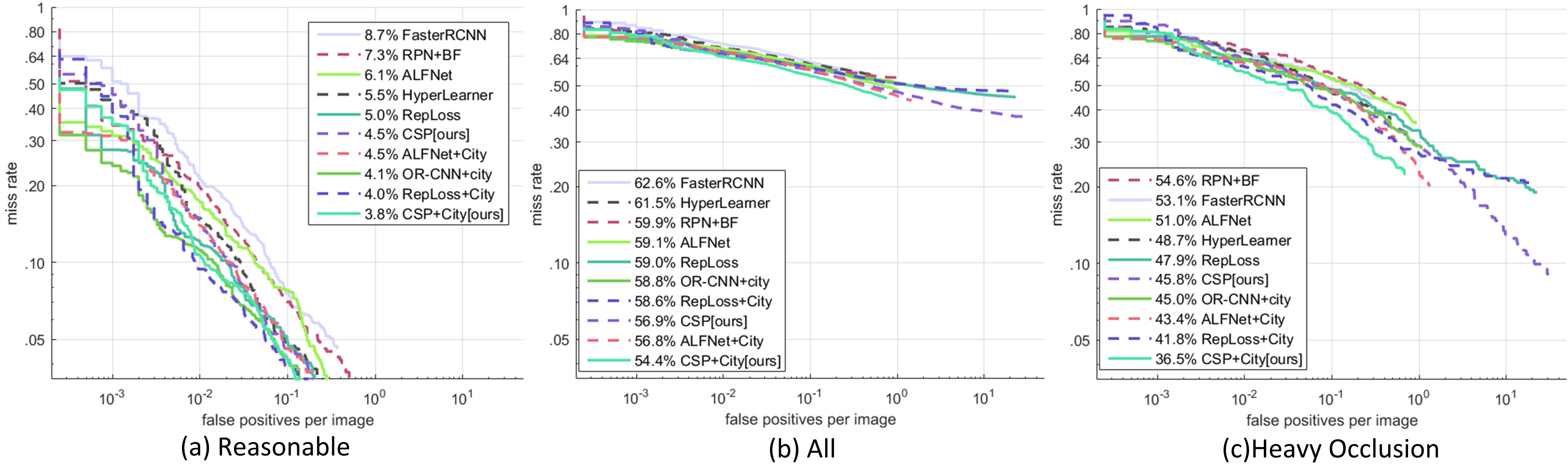}
\end{center}
   \caption{Comparisons with the state of the arts on Caltech using new annotations.}
\label{fig:cal}
\end{figure*}

\begin{table*}
\begin{center}
\resizebox{.999\textwidth}{!}{
\begin{tabular}{c|c|c|c c c|c c c|c}
\hline
Method & Backbone & Reasonable & Heavy & Partial & Bare & Small & Medium & Large & Test Time\\
\hline
\hline
FRCNN\cite{zhang2017citypersons} & VGG-16 & 15.4 & - & - & - & 25.6 & 7.2 & 7.9 & -\\
\hline
RetinaNet \cite{lin2017focal}  & ResNet-50 & 15.6 & 49.98 & - & - & - & - & - & -\\
\hline
CornerNet \cite{Law_2018_ECCV} & Hourglass-54 & 21.0 & 56.0 & - & - & - & - & - & -\\
\hline
FRCNN+Seg\cite{zhang2017citypersons} & VGG-16 & 14.8 & - & - & - & 22.6 & 6.7 & 8.0 & -\\
\hline
OR-CNN\cite{Zhang_2018_ECCV} & VGG-16 & 12.8 & 55.7 & 15.3 & \color{red}{6.7} & - & - & - & -\\
\hline
RepLoss\cite{wang2017repulsion} & ResNet-50 & 13.2 & 56.9 & 16.8 & 7.6 & - & - & -& -\\
\hline
TLL\cite{Song_2018_ECCV} & ResNet-50 & 15.5 & 53.6 & 17.2 & 10.0 & - & - & -& -\\
\hline
TLL+MRF\cite{Song_2018_ECCV} & ResNet-50 & 14.4 & 52.0 & 15.9 & 9.2 & - & - & -& -\\
\hline
ALFNet\cite{Liu_2018_ECCV} & ResNet-50 & 12.0 & 51.9 & 11.4 & 8.4 & 19.0 & 5.7 & 6.6 & 0.27s/img\\
\hline
CSP(w/o offset) & ResNet-50 & \color{green}{11.4} & \color{green}{49.9} & \color{green}{10.8} & 8.1 & \color{green}{18.2} & \color{green}{3.9} & \color{red}{6.0}& 0.33s/img\\
\hline
CSP(with offset) & ResNet-50 & \color{red}{11.0} & \color{red}{49.3} & \color{red}{10.4} & \color{green}{7.3} & \color{red}{16.0} & \color{red}{3.7} & \color{green}{6.5}& 0.33s/img\\
\hline
\end{tabular}
}
\end{center}
\caption{Comparison with the state of the arts on CityPersons\cite{zhang2017citypersons}. Results test on the original image size (1024x2048 pixels) are reported. {\color{red}{Red}} and {\color{green}{green}} indicate the best and second best performance.}
\label{table:cityval}
\end{table*}

\subsection{Comparison with the State of the Arts}
\subsubsection{Pedestrian Detection}
\textbf{Caltech.}
Extensive comparisons are conducted on three settings: Reasonable, All and Heavy Occlusion.
As shown in Fig. \ref{fig:cal}, CSP achieves $MR^{-2}$ of 4.5\% on the Reasonable setting, outperforming the best competitor (5.0 of RepLoss \cite{wang2017repulsion}) by 0.4\%. With the model initialized from CityPersons\cite{zhang2017citypersons}, CSP achieves a new state of the art of 3.8\%, compared to 4.0\% of RepLoss \cite{wang2017repulsion}. A mere  pre-training on CityPersons further boosts the performances because Caltech has lower person per images density compared to CityPersons \footnote{ CityPersons has roughy 6 persons/image whereas Caltech has 0.3 \cite{shao2018crowdhuman}}. It presents the superiority on detecting pedestrians of various scales and occlusion levels as demonstrated in Fig . \ref{fig:cal} (b). Moreover, Fig. \ref{fig:cal} (c) shows that CSP also performs very well for heavily occluded pedestrians, outperforming RepLoss \cite{wang2017repulsion} and OR-CNN \cite{Zhang_2018_ECCV} which are explicitly designed for occlusion cases.

\textbf{CityPersons.}
Table \ref{table:cityval} shows the comparison with previous state of the arts on CityPersons.
Following \cite{wang2017repulsion, zhang2017citypersons}, results on subsets with different occlusion levels and various scale ranges are also reported. It can be observed that CSP beats the competitors and performs fairly well on occlusion cases even without any specific occlusion-handling strategies \cite{wang2017repulsion,Zhang_2018_ECCV}.
On the Reasonable subset, CSP performs the best with a gain of 1.0\% $MR^{-2}$ upon the closest competitor (ALFNet \cite{Liu_2018_ECCV}), while the speed is comparable on the same running environment with 0.33 second per image of 1024x2048 pixels. Additionally, we also compare CSP with state of the arts (RetinaNet \cite{lin2017focal} and CornerNet \cite{Law_2018_ECCV}) in generic object detection, results presented in Table \ref{table:cityval} demonstrate the superority of CSP in pedestrian detection.

\textbf{CrowdHuman.} We further validated our approach on a recently collected large-scale general purpose (curated from web-crawling) person detection dataset, CrowdHuman \cite{shao2018crowdhuman}. Unlike CityPersons and Caltech, CrowdHuman dataset does not have a fixed aspect ratio. CrowdHuman benchmark is a more diverse and dense dataset than CityPerson and Caltech in terms of person per image and unique pedestrians. We compare with a single stage object detector RetinaNet \cite{lin2017focal} in Table \ref{table:crowdhuman}. CSP outperforms RetinaNet \cite{lin2017focal} on a general person detection benchmark by more than 1 $MR^{-2}$(\%), illustrating the robustness of the proposed approach.

\begin{table}
\begin{center}
\begin{tabular}{c|c}
\hline
Methods & {$MR^{-2}$(\%)}  \\
\hline
\hline
RetinaNet\cite{lin2017focal} & 63.3 \\
\hline
CSP[ours] & \textbf{62.1} \\
\hline
\end{tabular}
\end{center}
\caption{Comparison of CSP with another single stage object detector on general person detection dataset, CrowdHuman).}
\label{table:crowdhuman}
\end{table}

\begin{figure*}[t]
\begin{center}
\includegraphics[width=1.0\linewidth]{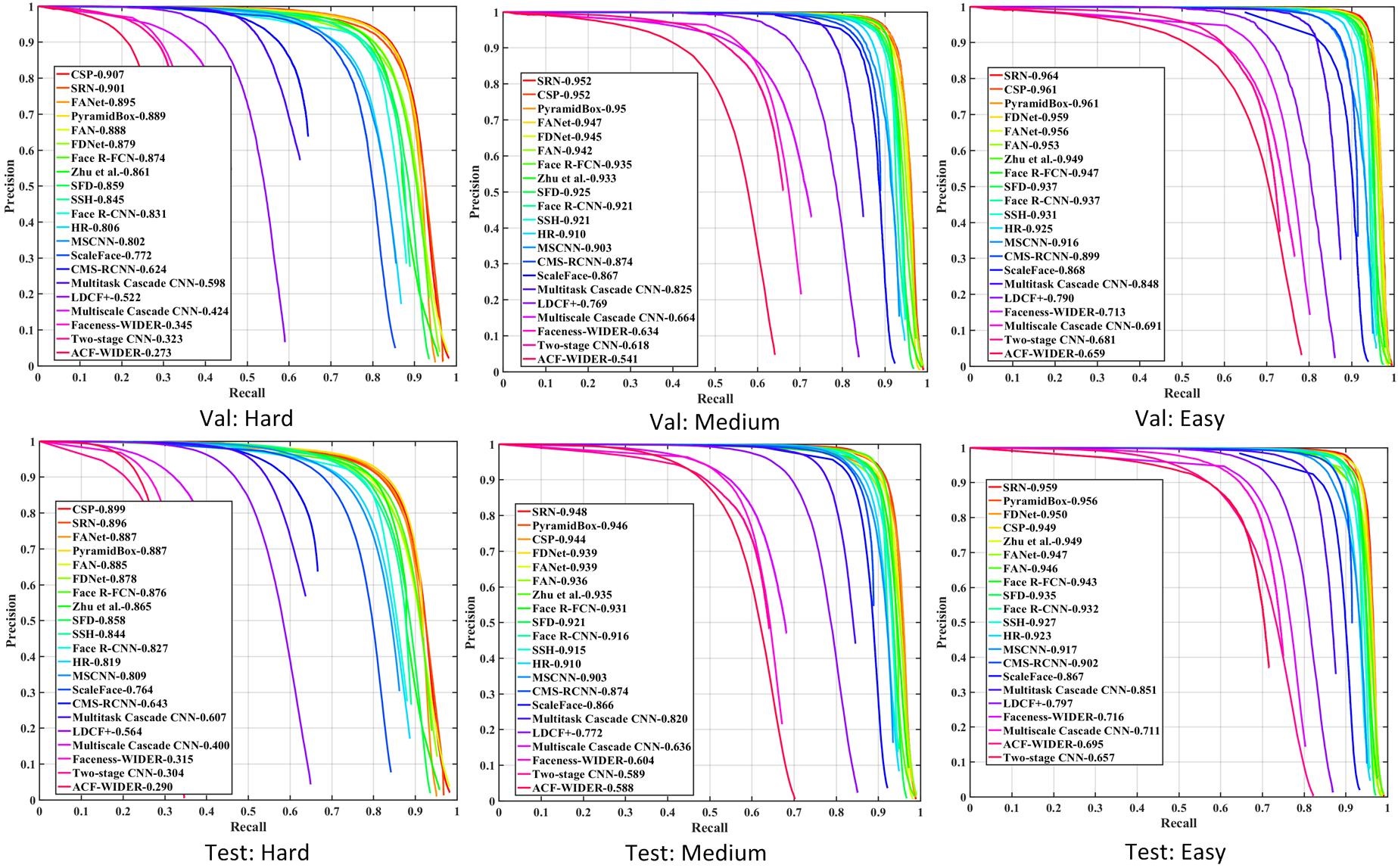}
\end{center}
   \caption{Precision-recall curves on WIDER FACE validation and testing subsets.}
\label{fig:wider}
\end{figure*}

\subsubsection{Face Detection}
\textbf{WiderFace}.
The model trained on the training subset of WiderFace are evaluated on both the validation and test subsets, and the multi-scale testing is also performed in a similar way as in \cite{Tang_2018_ECCV}. Comparisons with the state-of-the-art face detectors on WiderFace are shown in Fig. \ref{fig:wider}. It can be seen that the proposed CSP detector achieves competitive performance among the state-of-the-art face detectors across the three subsets, i.e. 90.7\% (Hard), 95.2\% (Medium) and 96.1\% (Easy) on validation subset, and 89.9\% (Hard), 94.4\% (Medium) and 94.9\% (Easy) on test subset. Note that most of these face detectors are anchor-box based. Therefore, the results indicate a superiority of the proposed CSP detector when complex default anchor-box design and anchor-box matching strategies are abandoned.

\subsection{Generalization ability of the proposed method}
To further demonstrate the generalization ability of the proposed CSP detector, we perform cross-dataset evaluation on two tasks, i.e. pedestrian detection and face detection. Specifically, models trained on the source dataset are directly tested on the target dataset without further finetuning. Furthermore, unlike humans, faces occur in different scale and aspect ratio, as shown in Table \ref{table:arcomp}. In Table \ref{table:arcomp}, we illustrate that in order to achieve high localization accuracy for faces, CSP needs to predict both height and width jointly, as oppose to simply predicting height (as in pedestrian detection).

\subsubsection{Cross-dataset evalutaion for Pedestrian Detection}
For pedestrian detection, we compare the proposed CSP detector with the state-of-the-art anchor-box based pedestrian detector (ALFNet \cite{Liu_2018_ECCV}). Both of the two detectors are trained on the CityPersons \cite{zhang2017citypersons} training subset and then are directly tested on the Caltech \cite{dollar2012pedestrian} test subset. For ALFNet \cite{Liu_2018_ECCV}, we use the source code and models provided by the authors\footnote{https://github.com/liuwei16/ALFNet}.
Results shown in Table \ref{table:1} are based on the reasonable setting, and the evaluation metric is log average Miss Rate (MR). It can be seen that the gap between the two detectors on the source dataset (CityPersons) is merely 1\%, but the gap on the target dataset (Caltech) increases to 5.9\%, which gives the evidence that the proposed CSP detector generalizes better to another dataset than the anchor-box based competitor, i.e. ALFNet \cite{Liu_2018_ECCV}.

\begin{table}
\begin{center}
\begin{tabular}{c|c|c}
\hline
Methods & CityPersons $\rightarrow$ CityPersons & CityPersons $\rightarrow$ Caltech\\
\hline
\hline
ALFNet\cite{Liu_2018_ECCV} & 12.0 & 17.8\\
\hline
CSP[ours] & \textbf{11.0} & \textbf{11.9}\\
\hline
\end{tabular}
\end{center}
\caption{Comparisons of the generalization ability for pedestrian detection (Evaluation metric: log average Miss Rate; the lower, the better).}
\label{table:1}
\end{table}

\begin{table}
\begin{center}
\begin{tabular}{c|c|c|c}
\hline
Dataset & FDDB \cite{fddbTech} & UCCS \cite{uccs} & DarkFace \cite{wei2018deep}\\
\hline
\hline
Num. of images & 2485 & 5232 & 6000\\
\hline
Num. of faces & 5171 & 11109 & 43849\\
\hline
Mean face size & 95x141 & 225x407 & 16x17\\
\hline
Mean image size & 377x399 & 5184x3456 & 1080x720\\
\hline
\end{tabular}
\end{center}
\caption{Statistics of three face detection datasets for cross-dataset evaluation.}
\label{table:2}
\end{table}

\begin{figure*}[t]
\begin{center}
\includegraphics[width=1.0\linewidth]{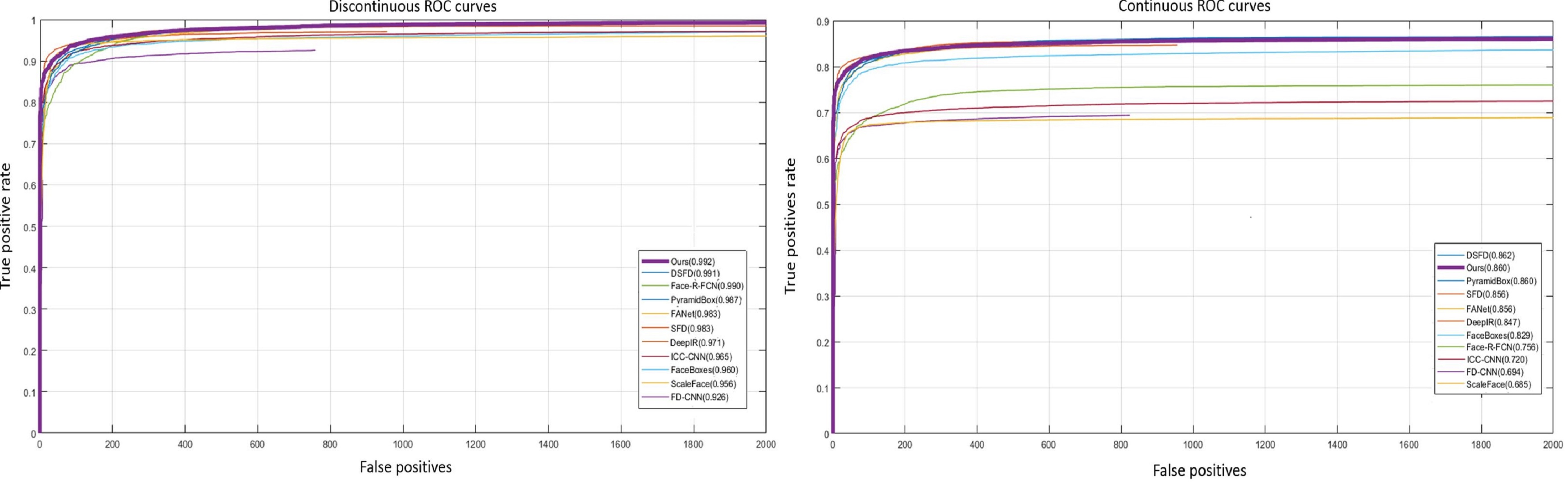}
\end{center}
   \caption{Comparisons of ROC results on the FDDB dataset.}
\label{fig:fddb}
\end{figure*}

\subsubsection{Cross-dataset evalutation for Face Detection}
For face detection, models trained on the WiderFace \cite{yang2016wider} training subset are directly tested on three other face detection datasets, i.e. FDDB \cite{fddbTech}, UCCS \cite{uccs} and DarkFace \cite{wei2018deep}. Detailed statistics about these three datasets for testing are listed in Table \ref{table:2}. It can be seen that these three datasets exhibit a large difference in the mean size of face objects.

FDDB \cite{fddbTech} is also a widely adopted face detection benchmark. Comparisons with other advanced face detectors on this benchmark are reported in Fig. \ref{fig:fddb}, results of other face detectors are from FDDB's official website\footnote{http://vis-www.cs.umass.edu/fddb/results.html}. As shown in Fig. \ref{fig:fddb}, the proposed CSP detector achieves competitive results on both discontinuous and continuous ROC curves, with the true positive rate of 99.2\% and 86.0\% when the number of false positives equals to 1000, while the results of the most recent anchor-box based face detector (DSFD \cite{li2018dsfd}) are 99.1\% and 86.2\%, respectively. Since face images of both FDDB and WiderFace are obtained from the Internet, they are basically similar. Therefore, both CSP and DSFD detectors trained on WiderFace perform quite good on FDDB, and there is little performance difference between them.

However, when evaluated on the other two quite different face datasets, UCCS \cite{uccs} and DarkFace \cite{wei2018deep}, it is interesting to see some difference.
UCCS \cite{uccs}, with the full name of UnConstrained College Students (UCCS), is a recently published dataset collected by a long-range high-resolution surveillance camera. The significant difference between UCCS \cite{uccs} and other face detection datasets is that the data are collected unconstrainedly in surveillance scenes. People walking on the sidewalk did not aware that they were being recorded. As the annotations of the test subset is publicly unavailable, results on the validation subset are reported.

DarkFace \cite{wei2018deep} is a recently published face detection dataset collected during nighttime, which exhibits an extreme light condition compared to other face detection datasets. The average size of the face objects in this dataset is merely 16x17 pixels. In the official website\footnote{https://flyywh.github.io/CVPRW2019LowLight/}, it totally released 6000 images, on which we test both models and report the results.

For UCCS \cite{uccs} and DarkFace \cite{wei2018deep}, we compare the proposed CSP with the state-of-the-art anchor-box based face detector (DSFD \cite{li2018dsfd}). For DSFD, we use the source code and models provided by the authors\footnote{https://github.com/TencentYoutuResearch/FaceDetection-DSFD}.
Results are given in Table \ref{table:3}, and the evaluation metric is the Average Precision (AP). As can be seen from Table \ref{table:3}, though the proposed CSP slightly underperforms DSFD on the WiderFace test subset, it achieves a significant gain over DSFD on these two cross-dataset evaluations, with 3.7\% and 2.1\% on UCCS \cite{uccs} and DarkFace \cite{wei2018deep}, respectively. Due to the substantial domain gaps between UCCS, DarkFace and WiderFace, both models trained on WiderFace perform unsurprisingly poor on UCCS and DarkFace, but the proposed CSP detector still outperforms the anchor-box based DSFD, which gives the evidence that CSP generalizes better to unknown domains than the anchor-box based competitor. It is possible that the default configurations of anchor-boxes in anchor-based detectors can not adapt to new scenes especially when the scales and aspect ratios of objects have a large difference as shown in Table \ref{table:2}. In contrast, the proposed detector simply predicts the centers and scales of objects without any considerations of priors of the objects in the dataset, thus shows a better generalization ability.

\begin{table}
\begin{center}
\resizebox{1.0\linewidth}{!}{
\begin{tabular}{c|c|c|c|c|c}
\hline
\multirow{2}{*}{Methods} & \multicolumn{3}{c|}{WiderFace $\rightarrow$ WiderFace} & \multirow{2}{*}{WiderFace $\rightarrow$ UCCS} & \multirow{2}{*}{WiderFace $\rightarrow$ DarkFace}\\
\cline{2-4} {} & Hard & Medium & Easy & {} & {}\\
\hline
\hline
DSFD \cite{li2018dsfd} & \textbf{90.0} & \textbf{95.3} & \textbf{96.0} & 7.6 & 25.9\\
\hline
CSP[ours] & 89.9 & 94.4 & 94.9 & \textbf{11.3} & \textbf{28.0}\\
\hline
\end{tabular}
}
\end{center}
\caption{Comparisons on generalization ability of face detectors (Evaluation metric: Average Precision (AP); the higher, the better).}
\label{table:3}

\end{table}

\begin{table}
\begin{center}
\begin{tabular}{c|c|c}
\hline
Disturbance (pixels) & $MR^{-2}(\%)$ & $\Delta MR^{-2}(\%)$\\
\hline
\hline
0 & 4.62 & -\\
\hline
[0, 4] & 5.68& $\downarrow1.06$\\
\hline
[0, 8] & 8.59& $\downarrow3.97$\\
\hline
\end{tabular}
\end{center}
\caption{Performance drop with disturbances of the centers.}
\label{table:disturb}
\end{table}

\begin{table}
\begin{center}
\begin{tabular}{c|c|c}
\hline
Method & CSP H + W & CSP H \\
\hline
\hline
Easy & 0.961 & 0.903\\
\hline
Medium & 0.952 & 0.898\\
\hline
Hard & 0.907 & 0.840 \\
\hline
\end{tabular}
\end{center}
\caption{CSP with H+W vs. CSP with H only.}
\label{table:arcomp}
\end{table}

\subsection{Further discussions}
Note that CSP only requires object centers and scales for training, though generating them from bounding box or central line annotations is more feasible since centers are not always easy to annotate. Besides, the model may be puzzled on ambiguous centers during training. To demonstrate this, we also conduct an ablative experiment on Caltech, in which object centers are randomly disturbed in the range of [0,4] and [0,8] pixels during training.
Results in Table \ref{table:disturb} show that performance drops with increasing annotation noise. For Caltech, we also apply the original annotations but with inferior performance to another anchor-free detector, TLL \cite{Song_2018_ECCV}. A possible reason is that TLL includes a series of post-processing strategies in keypoint pairing. As evaluations of TLL on Caltech with new annotations\cite{zhang2016far} are not reported in \cite{Song_2018_ECCV}, comparison to TLL is given in Table \ref{table:cityval} on the CityPersons, which shows the superiority of CSP. Therefore, the proposed method may be limited for annotations with ambiguous centers, e.g. the traditional pedestrian bounding box annotations affected by limbs. In view of this, applying CSP to generic object detection requires further improvement.

\begin{figure*}[t]
\begin{center}
\includegraphics[width=1.0\linewidth]{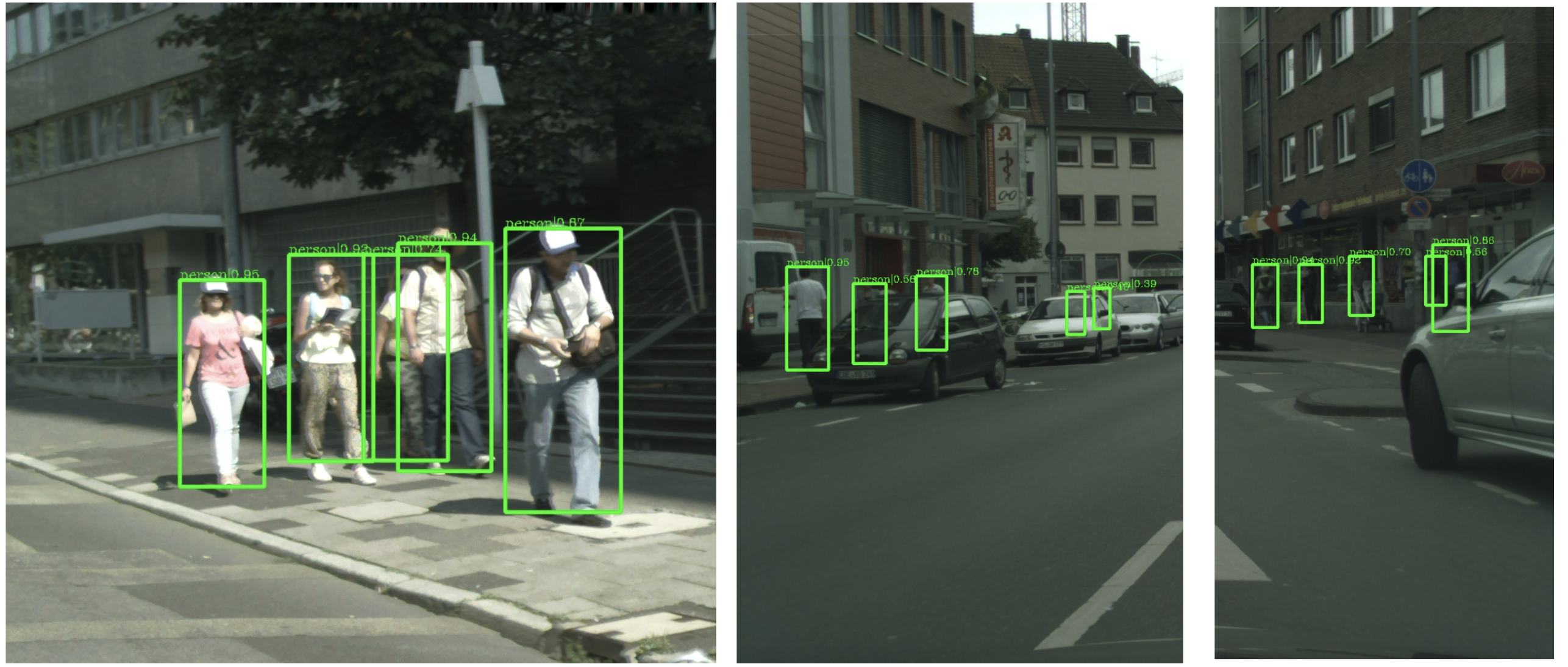}
\end{center}
   \caption{Different type of challenges present in pedestrian detection such as person-over-person occlusion (left), occlusion due to cars and objects (middle) and low illumination (right). CSP is robust enough to handle such challenging scenarios}
\label{fig:qual}
\end{figure*}

When compared with anchor-box based methods, the advantage of CSP lies in two aspects. Firstly, CSP does not require tedious configurations on anchor-boxes specifically for each dataset. Secondly, anchor-box based methods detect objects by overall classifications of each anchor-box where background information and occlusions are also included and will confuse the detector's training. However, CSP overcomes this drawback by scanning for pedestrian centers instead of anchor-boxes in an image, thus is more robust to occluded objects as shown in Fig. \ref{fig:qual}.

\section{Conclusion}

Deviating from traditional paradigms for feature detection, in this work, we argue in the favor of posing pedestrian detection as a high-level semantic feature detection task through straightforward convolutions for center and scale predictions. This enables a complete anchor.free settings and is also free from complex post-processing strategies as in recent keypoint-pairing based detectors. Consequently, the proposed CSP detector achieves the new state-of-the-art performance on two challenging pedestrian detection benchmarks, namely CityPersons and Caltech. Due to the generic architecture of the CSP detector, we further evaluate it for face detection on the most popular face detection benchmark, i.e. WiderFace. The comparable performance to other advanced anchor-box based face detectors also shows the proposed CSP detector's competitiveness. Besides, experiments on cross-dataset evaluation for both pedestrian detection and face detection further demonstrate CSP's superior generalization ability over anchor-box based detectors. For future possibilities, it is interesting to further explore CSP's capability in general object detection. Given its superiority on cross-dataset evaluation, it is also interesting to see CSP's potential when domain adaptation techniques are further explored.


\bibliography{csp}

\end{document}